\documentclass{article}
\usepackage{spconf,amsmath,graphicx}
\usepackage{amsfonts,wrapfig}
\newcommand*{\affmark}[1][*]{\textsuperscript{#1}}

\title{Weighted Linear Discriminant Analysis \\based on Class Saliency Information}
\name{Lei Xu\affmark[1], Alexandros Iosifidis\affmark[2] and Moncef Gabbouj\affmark[1]}
\address{\affmark[1]Laboratory of Signal Processing, Tampere University of Technology, Tampere, Finland \\
\affmark[2] Department of Engineering, Electrical \& Computer Engineering, Aarhus University, Aarhus, Denmark}

\begin{document}
%
\maketitle
\begin{abstract}
In this paper, we propose a new variant of Linear Discriminant Analysis to overcome underlying drawbacks of traditional LDA and other LDA variants targeting problems involving imbalanced classes. Traditional LDA sets assumptions related to Gaussian class distribution and neglects influence of outlier classes, that might hurt in performance. We exploit intuitions coming from a probabilistic interpretation of visual saliency estimation in order to define saliency of a class in multi-class setting. Such information is then used to redefine the between-class and within-class scatters in a more robust manner. Compared to traditional LDA and other weight-based LDA variants, the proposed method has shown certain improvements on facial image classification problems in publicly available datasets.
\end{abstract}
\begin{keywords}
Visual saliency estimation, Fisher's discriminant criterion
\end{keywords}
\section{Introduction}
\label{sec:intro}

Linear Discriminant Analysis (LDA), as a traditional statistical machine learning technique, has been employed for several classification tasks, such as human action recognition \cite{Iosifidis2012Multi-viewAnalysis}, \cite{Iosifidis2014RegularizedRecognition} and person identification \cite{Iosifidis2012Activity-basedLearning}, due to its effectiveness in reducing dimensions and extracting discriminative features. In a classification task, LDA is used to define an optimal projection by means of Fisher criterion optimization. Despite the widespread application of traditional LDA, its performance is affected by several issues related to its underlying assumptions. Traditional LDA represents each class with the corresponding class mean and discriminates between classes based on the scatters of these class representations with respect to the total data mean. Such a class discrimination definition may cause large overlaps of neighboring classes \cite{Yu2002ARecognition}, and receive a sub-optimal result, since an outlier class being far from the others dominates the solution \cite{Yu2002ARecognition}. Furthermore, in traditional LDA all classes equally contribute to the within-class scatter definition \cite{Tang2004} based on the assumption of the same Gaussian distribution for all classes. This assumption overemphasizes well-separated outlier classes, which should have lower contribution in the overall within-class scatter definition. A method that automatically determines optimized class representations for LDA-based projections was proposed in \cite{Iosifidis2013OnAnalysis}, \cite{Iosifidis2014KernelAnalysis}; however, it also suffers from the class imbalance problems discussed above. In order to overcome aforementioned drawbacks of traditional LDA, extensions imposing weighting strategies for the definition of the within-class and between-class scatters have been proposed in \cite{Ahmed2012FaceAlgorithm}, \cite{Tang2005LinearLDA}, \cite{Jarchi2008ALDA}, \cite{Loog2001MulticlassCriteria}, \cite{Li2009NonparametricRecognition}. In these methods, the weighting factors incorporated to the scatter matrices definitions are based on class statistics, e.g. class cardinality, and class representation is still assumed to be the class mean. 

A novel extension of LDA that exploits intuitions from saliency \cite{Aytekin2016RecentDetection} is proposed in this paper. A probabilistic criterion is formulated in order to express the samples around boundary within its original class following a probabilistic saliency estimation framework \cite{Aytekin2016}. Such a definition is naturally expressed by graph notation, in which several types of graphs can be exploited. Both fully connected and $k$-NN graphs are considered. After defining the probability of each sample belonging to its corresponding class, this information is used to define new class representations, as well as new within-class and between-class scatters. Compared to traditional LDA and its weighted variants, the proposed Saliency-based weighted LDA ($SwLDA$) has shown enhanced performance on facial image classification problems. 

The remainder of this paper is structured as follows. In Section 2, we briefly present related works. In Section 3, we rigorously derive the proposed $SwLDA$ method on the basis of various weighted LDA methods and saliency estimation. Experimental results on publicly available facial image datasets are provided in Section 4, and Section 5 concludes this work.

\section{related work}
\label{sec:reWork}

In this section, first we briefly describe original LDA and two of its weighted variants, which have been proposed in order to overcome shortcomings of LDA related to class imbalance problems. Later, visual saliency estimation based on the recently proposed probabilistic interpretation \cite{Aytekin2016} is presented. 

In the following, we assume that each training sample is represented by a vector $\mathbf{x}_i \in \mathbb{R}^D$ and is followed by a class label $y_i \in \{1,\dots,C\}$. A set of training vectors $\mathbf{x}_i, \:i=1,\dots,N$ are used in order to define a linear projection from the input space $\mathbb{R}^D$ to a discriminant subspace $\mathbb{R}^d$ such that the representation of the $i$-th sample is given by $\boldsymbol{z}_i = \mathbf{W}^T \mathbf{x}_i$, where $\mathbf{W}\in\mathbb{R}^{D \times d}$ is the projection matrix to be learned by optimizing class discrimination criteria.

\subsection{Linear Discriminant Analysis}   
LDA defines the optimal data projection matrix $\mathbf{W}$ by maximizing the following criterion
\begin{equation} \label{eq:1}
\scalebox{0.9}[0.9]{$J(\mathbf{W}) = \underset{\mathbf{W}}{max} \:\: \frac{tr(\mathbf{W}^T \mathbf{S}_B \mathbf{W})}{tr(\mathbf{W}^T \mathbf{S}_W \mathbf{W})}$},
\end{equation}
where $\mathbf{S}_W$, $\mathbf{S}_B$ are within-class and between-class scatter matrices respectively, and defined as follows:
\begin{equation} \label{eq:2}
\scalebox{0.9}[0.9]{$\mathbf{S}_W = \sum_{c=1}^{C}\sum_{\mathbf{x}_i, \alpha_i^c}(\mathbf{x}_i - \mbox{\boldmath$\mu$}_c)(\mathbf{x}_i - \mbox{\boldmath$\mu$}_c)^T$},
\end{equation}
\begin{equation} \label{eq:3}
\scalebox{0.9}[0.9]{$\mathbf{S}_B = \sum_{c=1}^{C} N_c(\mbox{\boldmath$\mu$}_c - \mbox{\boldmath$\mu$})(\mbox{\boldmath$\mu$}_c - \mbox{\boldmath$\mu$})^T$}.
\end{equation}
In the above, $\alpha_i^c$ is an index denoting whether sample $i$ belongs to class $c$, i.e. $\alpha_i^c = 1$ if $y_i = c$ and $\alpha_i^c = 0$ otherwise. $N_c$ denotes the cardinality of class $c$, i.e. $N_c = \sum_{i=1}^N \alpha_i^c$ and $\mbox{\boldmath$\mu$}_c$ denotes the mean vector of class $c$, i.e. $\mbox{\boldmath$\mu$}_c = \frac{1}{N_c} \sum_{\mathbf{x}_i, \alpha_i^c = 1} \mathbf{x}_i$. $\mbox{\boldmath$\mu$}$ is the total mean vector $\mbox{\boldmath$\mu$} = \frac{1}{N} \sum_{i=1}^N \mathbf{x}_i$.

The optimal projection matrix $\mathbf{W}$ is obtained through applying eigenvalue decomposition of the matrix $\mathbf{S} = \mathbf{S}_{W}^{-1}\mathbf{S}_{B}$ and keeping the eigenvectors corresponding to the largest (up to $C-1$ in total) eigenvalues.

\subsection{Weighted LDA Variants}
\label{sec:wlda}
Weighted versions of LDA aim at scaling the contribution of each class based on their influences on projection, by defining appropriate weights. In \cite{Loog2001MulticlassCriteria}, between-class scatter matrix is redefined for enhancing robustness in multi-class problems, as follows:
\begin{equation} \label{eq:4}
\scalebox{0.9}[0.9]{$\mathbf{S}_b = \sum_{c=1}^{C-1}\sum_{j=c+1}^{C}L_{cj} p_c p_j (\mbox{\boldmath$\mu$}_c - \mbox{\boldmath$\mu$}_j)(\mbox{\boldmath$\mu$}_c - \mbox{\boldmath$\mu$}_j)^T$},
\end{equation}
where $p_c$, $p_j$ denote the prior probability of class $c$, class $j$, respectively. $L_{cj}$ expresses the dissimilarity between class $c$ and class $j$, using a distance function in the Euclidean (or a Mahalanobis) space. In order to reduce the influence of outlier classes, an outlier-class-resistant weighted LDA method is proposed in this work \cite{Tang2005LinearLDA} based on Loog's work \cite{Loog2001MulticlassCriteria}. They express the between-class scatter using (\ref{eq:4}) and a new within-class scatter definition is proposed as follows:
\begin{equation} \label{eq:5}
\scalebox{0.9}[0.9]{$\mathbf{S}_w = \sum_{c=1}^{C} \sum_{k=1}^{N_c} p_c r_c (\mathbf{x}_k - \mbox{\boldmath$\mu$}_c) (\mathbf{x}_k - \mbox{\boldmath$\mu$}_c)^T$},
\end{equation}
where $r_c=\sum_{i \neq c}\frac{1}{L_{ic}}$ is a relevance-weight between class $c$ and class $i$, reducing attention to outlier classes.

Another version of weighted LDA aiming at alleviating the influence of outlier class is proposed in \cite{Jarchi2008ALDA}. They define the between-class scatter and within-class scatter as follows:
\begin{equation} \label{eq:7}
\scalebox{0.9}[0.9]{$\mathbf{S}_{b} = \sum_{c=1}^{C-1} \sum_{j=c+1}^{C} n_c n_j w_1(\Delta_{cj})(\mbox{\boldmath$\mu$}_c - \mbox{\boldmath$\mu$}_j) (\mbox{\boldmath$\mu$}_c - \mbox{\boldmath$\mu$}_j)^T$},
\end{equation}
\begin{equation}
\scalebox{0.9}[0.9]{$\mathbf{S}_w = \sum_{c=1}^{C} \sum_{k=1}^{N_c} p_c w_2(\Delta_{c:}) (\mathbf{x}_k - \mbox{\boldmath$\mu$}_c) (\mathbf{x}_k - \mbox{\boldmath$\mu$}_c)^T$},
\end{equation}
where $n_c$, $n_j$ are the number of samples for class $c$ and class $j$, in addition, $w_1(\Delta_{cj})$ and $w_2(\Delta_{c:})$ are defined as $\frac{1}{\Delta_{cj}}$ and $\frac{1}{\sum_{j \neq c} \Delta_{c j}}$, respectively. $\Delta_{cj}$ is the Fisher's discriminant criterion in the discriminant space determined through applying LDA using the between-class scatter matrix $\mathbf{S}_B$ and the total scatter matrix $\mathbf{S}_T = \mathbf{S}_B + \mathbf{S}_W$, i.e.:
\begin{equation} \label{eq:9}
\scalebox{1}[1]{$\mathbf{w}^* = \underset{\mathbf{w}}{argmax} \{ \frac{\mathbf{w}^T \mathbf{S}_B \mathbf{w}}{\mathbf{w}^T \mathbf{S}_T \mathbf{w}}\} = \mathbf{S}_T^{-1} (\mbox{\boldmath$\mu$}_c - \mbox{\boldmath$\mu$}_j)$},
\end{equation}
\begin{equation} \label{eq:10}
\scalebox{1}[1]{$\Delta_{cj} = \frac{\mathbf{w}^{*T} \mathbf{S}_B \mathbf{w}^*}{\mathbf{w}^{*T} \mathbf{S}_T \mathbf{w}*}$}.
\end{equation}
\indent
Using the above definition of $\Delta_{cj}$, in the case where a class is well separated from all others, a smaller value of $w(\Delta_{cj})$ will be used, reducing the influence of that class on the result. Once the new $\mathbf{S}_w$ and $\mathbf{S}_t$ ($\mathbf{S}_t$ = $\mathbf{S}_w$ + $\mathbf{S}_b$) are obtained, the final projection matrix $\mathbf{W}$ can be determined by optimizing the following Fisher's discriminant criterion:
\begin{equation} \label{eq:12}
\scalebox{0.9}[0.9]{$J(\mathbf{W}) = \underset{\mathbf{W}}{argmax} \frac{tr(\mathbf{W}^T \mathbf{S}_b \mathbf{W})}{tr(\mathbf{W}^T \mathbf{S}_t \mathbf{W})}$}.
\end{equation}

\subsection{Visual Saliency Estimation}
\label{sec:saliency}
Visual saliency estimation has gained attention during the last decade, since it can be applied as a pre-processing step for higher level Computer Vision tasks. Recently, Aytekin et al. formulated the salient object segmentation problem based on probabilistic interpretation. Specifically, they defined a probability mass function $P(x)$ encoding the probability that an image region (in the sense of pixel, super-pixel or patch) to depict a salient region. Estimation of $P(x)$ is formulated as an optimization problem enforcing similar regions to have similar probabilities, while any prior information regarding saliency (defined based on the location of each region in the image lattice) can be exploited. This joint optimization is expressed as:
\begin{gather}
\scalebox{0.6}[0.9]{$\underset{r(x)}{argmin}\bigg(\sum_i(P(x=x_i))^2v_i + \frac{1}{2}\bigg(\sum_{i,j}\Big(\big(P(x = x_i)\big)^2-P(x=x_i)P(x=x_j) \Big)w_{i,j}\bigg) \bigg)$} \label{eq:13} \\
s.t. \scalebox{0.9}[0.9]{$\quad \sum_i P(x = x_i) = 1$}, \nonumber
\end{gather}
where $v_i \ge 0$ denotes prior information for region $i$ by non-negative values and $w_{ij}$ expresses the similarity of regions $i$ and $j$. The optimization problem in (\ref{eq:13}) can be expressed using a matrix notation as follows:
\begin{gather}
\scalebox{0.9}[0.9]{$\mathbf{p^*} = \underset{p}{argmin}(\mathbf{p^T H p})$}, \label{eq:14}\\
\scalebox{0.8}[0.8]{$\mathbf{H} = \mathbf{D} - \mathbf{W} + \mathbf{V}, \label{eq:15} $}\\
s.t.\quad \scalebox{0.9}[0.9]{$\mathbf{p^T1} = 1$},\nonumber
\end{gather}
\noindent
where $\mathbf{p}$ is a vector having elements $p_i = P(x = x_i)$ corresponding to the probability of each region to be salient. $\mathbf{W}$ is the affinity matrix of a graph having as vertices for the region representations and $\mathbf{D}$ is the corresponding diagonal matrix having elements equal to $D_{ii} = \sum_j \mathbf{W}_{ij}$. $\mathbf{V}$ is a diagonal matrix having elements $[\mathbf{V}]_{ii} = v_i$. In visual saliency, the element $V_{ii}$ expresses the a priori knowledge that an image location belongs to background, that is introduced by the user.

As has been shown in \cite{Aytekin2016}, the optimization problem in (\ref{eq:14}) has a global optimum given by: $\mathbf{p}_{pse}^* = \mathbf{H}^{-1} \mathbf{1}$. Interestingly, the above solution is equivalent to an one-class classification model, making a connection between salient object segmentation and one-class classification problems. In the following, we will use this connection in order to derive a new definition for class-representation and scatter matrices calculation in LDA.

\section{Saliency-based weighted Linear Discriminant Analysis}
\label{sec:SWLDA}
This section describes in detail the proposed weighted versions of LDA. We define the contribution of each sample to the corresponding class, and then new class representations and scatter matrices are proposed accordingly. We start by describing the proposed sample weights.

\subsection{Sample Weights and Class Representation}
\label{sample}
Weighted LDA variants represent each class with the corresponding mean vector and define weights based on pair-wise class distances to address the outlier class problem. Such mutation yields a certain improvement over traditional LDA. Nevertheless, it neglects the influences of outlier samples within each class \cite{Li2009NonparametricRecognition}, which may affect the classification result greatly. This is due to the fact that all class samples equally contribute to the definition of the class representation and scatter matrix calculation.

In our work, we determine the contribution of each sample based on its \textit{class saliency information}. We define the class saliency information of a sample $\mathbf{x}_i$ based on its probability to belong to its true class $y_i$. In order to do so, we calculate the probability mass function $P_c(x)$ of each class $c$ independently following the probabilistic saliency estimation (PSE) in \cite{Aytekin2016}. That is, for each class $c$, we form the corresponding graph $\mathcal{G}_C = \{\mathbf{X}_c, \mathbf{W}_c\}$, where $\mathbf{X}_c \in \mathbb{R}^{D \times N_c}$ is a matrix formed by the samples belonging to class $c$ and $\mathbf{W}_c \in \mathbb{R}^{N_c \times N_c}$ is the graph weight matrix expressing the similarity between the class samples. Any type of graph can be used to this end. In our experiments we have used fully connected and the $k$-NN graphs, using the heat kernel function:
\begin{equation} \label{eq:20}
\scalebox{0.9}[0.9]{$W_{ij} = \exp\left(-\frac{\|\mathbf{x}_i - \mathbf{x}_j\|}{ 2 \sigma^2}\right)$},
\end{equation}
where the value of $\sigma$ is set equal to the mean Euclidean distance between the class samples, which is the natural scaling factor for each class.

We define a priori saliency information as misclassification-based probability for the class data to be set in the diagonal elements of the matrix $\mathbf{V}_c$. Misclassification-based probability assumes that a sample is less probable to have high saliency information if it is closer to another class, when compared to its true class. In this case, the elements of $\mathbf{V}_c$ are set equal to:
        \begin{eqnarray}\label{eq:19}
          \begin{array}{r@{}l}
            V_{c,ii} =
          \end{array}
            \left\lbrace
          \begin{array}{ll}
            0,  &if \:\:d_{c,i}^c < \underset{k\neq c}{min} \:\:d_{c,i}^k , \\
            \frac{ d_{c,i}^c }{ \underset{k\neq c}{min} \:\:d_{c,i}^k },      & otherwise,
          \end{array}
          \right.
        \end{eqnarray}
where $d_{c,i}^k = \|\mathbf{x}_{c,i} - \mbox{\boldmath$\mu$}_k\|^2_2$. In this case, a sample which is close to another class is assigned to low saliency information, even if it may be close to the center of its class.

After having defined the matrices $\mathbf{W}_c$ and $\mathbf{V}_c$, the probability of each sample $\mathbf{x}_{c,i}$ to belong to class $c$ is given by:
$\mathbf{p}_{c} = \mathbf{H}_c^{-1} \mathbf{1}$, where $\mathbf{H}_c = \mathbf{D}_c - \mathbf{W}_c + \mathbf{V}_c$ and $\mathbf{D}_{c,ii} = \sum_j \mathbf{W}_{c,ij}$. Having obtained $\mathbf{p}_c \in \mathbb{R}^{N_c}, \:c=1,\dots,C$, we define a new class representation as $\mathbf{m}_c = \mathbf{X}_c \mathbf{p}_c$.

\begin{table*}[!ht]
\caption{Classification accuracy of proposed $SwLDA$ }\label{result1}
\vspace{1.0mm}
\scalebox{0.8}{
\begin{tabular}{||c|c|c|c|c|c|c|c|c|c|c|c|c||}
\hline
Dataset& \multicolumn{2}{c|}{BU} & \multicolumn{2}{c|}{KANADE} & \multicolumn{2}{c|}{JAFFE} & \multicolumn{2}{c|}{ORL} & \multicolumn{2}{c|}{YALE} & \multicolumn{2}{c||}{AR} \\
\hline
$K$&1&\scalebox{0.6}[1]{$\min(5,0.1*N_{c})$}&1&\scalebox{0.6}[1]{$\min(5,0.1*N_{c})$}&1&\scalebox{0.6}[1]{$\min(5,0.1*N_{c})$}&1&\scalebox{0.6}[1]{$\min(5,0.1*N_{c})$}&1&\scalebox{0.6}[1]{$\min(5,0.1*N_{c})$}&1&\scalebox{0.6}[1]{$\min(5,0.1*N_{c})$}\\
\hline
$SwLDA_{11}$ &0.5714&0.5714&0.6816&0.6939&0.5619&0.5762&0.9700&0.9700&$\boldsymbol{0.9597}$&0.9564&$\boldsymbol{0.9696}$&$\boldsymbol{0.9696}$\\
\hline
$SwLDA_{21}$ &0.5714&0.5686&0.6816&0.6816&0.5619&0.5762&0.9700&0.9700&$\boldsymbol{0.9597}$&0.9568&$\boldsymbol{0.9696}$&$\boldsymbol{0.9696}$\\
\hline
$SwLDA_{31}$ &0.5886&0.5829&0.6776&0.6776&0.5524&0.5762&$\boldsymbol{0.9850}$&$\boldsymbol{0.9850}$&$\boldsymbol{0.9597}$&0.9556&$\boldsymbol{0.9696}$&0.9692\\
\hline
$SwLDA_{41}$
&0.6500&0.6529&0.7020&0.6980&$\boldsymbol{0.5905}$&0.5857&$\boldsymbol{0.9850}$&$\boldsymbol{0.9850}$&$\boldsymbol{0.9597}$&0.9568&$\boldsymbol{0.9696}$&$\boldsymbol{0.9696}$\\
\hline
$SwLDA_{12}$  &0.5800&0.5814&0.6816&0.6816&0.5667&0.5667&$\boldsymbol{0.9850}$&$\boldsymbol{0.9850}$&0.9589&0.9564&0.9692&0.9688\\
\hline
$SwLDA_{22}$ &0.5800&0.5814&0.6816&0.6816&0.5667&0.5571&$\boldsymbol{0.9850}$&$\boldsymbol{0.9850}$&0.9589&0.9572&0.9684&0.9684\\
\hline
$SwLDA_{32}$ &0.6243&0.6200&0.6776&0.6776&0.5286&0.5238&0.9600&0.9600&0.9589&0.9572&0.9684&0.9684\\
\hline
$SwLDA_{42}$  &$\boldsymbol{0.6786}$&0.6743&$\boldsymbol{0.7224}$&0.7184&0.5476&0.5524&0.9450&0.9450&0.9593&0.9572&$\boldsymbol{0.9696}$&0.9692\\
\hline
\end{tabular}}
\end{table*}
\vspace{0cm}
\subsection{Scatter Matrices Definition}
\label{scatter}
By exploiting class-specific saliency information described above, we can define within-class scatter matrix in two different ways. The first one is to incorporate $\mathbf{p}_c$ in $\mathbf{S}_w$ as: 
\begin{equation} \label{eq:21}
\scalebox{0.9}[0.9]{$\mathbf{S}_{w}^{(1)} = \sum_{c=1}^{C} \sum_{j=1}^{N_c} p_{c,j} (\mathbf{x}_{c,j} - \mbox{\boldmath$\mu$}_c)(\mathbf{x}_{c,j} - \mbox{\boldmath$\mu$}_c)^T$},
\end{equation}
where $\mathbf{x}_{c,j}$ denotes $j$-th sample in class $c$, $p_{c,j}$ is saliency score for $j$-th sample in class $c$. The other one is inspired by relevance weighted LDA mentioned in section 2, as:
\begin{equation} \label{eq:22}
\scalebox{0.9}[0.9]{$\mathbf{S}_{w}^{(2)} = \sum_{c=1}^{C} \sum_{j=1}^{N_c} p_{c,j} r_{c} (\mathbf{x}_{c,j} - \mbox{\boldmath$\mu$}_c) (\mathbf{x}_{c,j} - \mbox{\boldmath$\mu$}_c)^T$}.
\end{equation}
Here $r_{c}=\sum_{i \neq c}\frac{1}{{L}_{i c}}$ is a relevance-weight, where ${L}_{ic}$ is defined based on the Euclidean distance between pairwise mean vectors of class $i$ and class $c$, as (\ref{eq:23}):
\begin{equation} \label{eq:23}
\scalebox{0.9}[0.9]{${L}_{i c} = \sqrt[]{(\mbox{\boldmath$\mu$}_i - \mbox{\boldmath$\mu$}_c)^T(\mbox{\boldmath$\mu$}_i - \mbox{\boldmath$\mu$}_c)}$}.
\end{equation}

Definitions of between-class scatter matrix in aforementioned LDA methods simply maximize either the variations between each class mean vector and the total mean vector, or the variations between class pairs. Here, we propose four types of between-class scatter matrices, which are not only based on the aforementioned definition of $\mathbf{S}_b$, but also capture the structure inside each class. The first definition is the same as (\ref{eq:3}):
\begin{equation} 
\scalebox{0.9}[0.9]{$\mathbf{S}_{b}^{(1)} = \sum_{c=1}^{C} N_c(\mbox{\boldmath$\mu$}_c - \mbox{\boldmath$\mu$})(\mbox{\boldmath$\mu$}_c - \mbox{\boldmath$\mu$})^T$}.
\end{equation}
\indent 
The second one uses saliency scores $\mathbf{p}_c$, when generating new class representations, as follows:
\begin{gather}
\scalebox{0.9}[0.9]{$\hat{\mbox{\boldmath$\mu$}}_c=\mathbf{X}_{c}\mathbf{p}_c$}, \label{eq:24} \\
\scalebox{0.9}[0.9]{$\mathbf{S}_{b}^{(2)} = \sum_{c=1}^C (\hat{\mbox{\boldmath$\mu$}}_c - \mbox{\boldmath$\mu$})(\hat{\mbox{\boldmath$\mu$}}_c - \mbox{\boldmath$\mu$})^T$}, \label{eq:25}
\end{gather}
where $\mathbf{X}_{c}$ contains all samples in class $c$, $\hat{\boldsymbol{\mu}}_c$ is the new class representation or weighted center of class $c$. The third definition extends (\ref{eq:25}) to exploit the relationships between pairs of new class representation for each class, as follows:
\begin{equation} \label{eq:26}
\scalebox{0.9}[0.9]{$\mathbf{S}_{b}^{(3)} = \sum_{c_1=1}^C \sum_{c_2=1}^C(\hat{\mbox{\boldmath$\mu$}}_{c_1} - \hat{\mbox{\boldmath$\mu$}}_{c_2})(\hat{\mbox{\boldmath$\mu$}}_{c_1} - \hat{\mbox{\boldmath$\mu$}}_{c_2})^T$}.
\end{equation}
\indent
The last definition, $\mathbf{S}_{b}^{(4)}$, intends to maximize discrimination between every sample in one class with other new class representations, meanwhile takes into account of each sample's saliency scores, as follows:
\begin{equation} \label{eq:27}
\scalebox{0.9}[0.9]{$\mathbf{S}_{b}^{(4)} = \sum_{c_1=1}^C \sum_{\substack{c_2=1,\\ c_2 \neq c_1}}^C \sum_{j=1}^{N_{c_1}} p_{c_1,j} (\mathbf{x}_{c_1,j} - \hat{\mbox{\boldmath$\mu$}}_{c_2})  (\mathbf{x}_{c_1,j} - \hat{\mbox{\boldmath$\mu$}}_{c_2})^T$},
\end{equation}
where $N_{c_1}$ is the cardinality of class $c_1$.

\subsection{Discriminant Criterion}
\label{Slda}
Using the above described scatter matrices, several optimization criteria can be formed as follows:
\begin{equation} \label{eq:28}
\scalebox{0.9}[0.9]{$J(\mathbf{W}) = \underset{\mathbf{W}}{argmax} \frac{tr(\mathbf{W}^T \mathbf{S}_{b}^{(i)} \mathbf{W})}{tr(\mathbf{W}^T \mathbf{S}_{t}^{(i j)} \mathbf{W})}$},
\end{equation}
where $\mathbf{S}_{t}^{(ij)}$ = $\mathbf{S}_{w}^{(j)}$ + $\mathbf{S}_{b}^{(i)}$, $i \in \{ 1, 2, 3, 4\} $ and $j \in \{ 1, 2 \} $. After obtaining projection matrix $\mathbf{W}$ by eigenvalue decomposition, we map corresponding class representations and test samples by the optimal $\mathbf{W}$, and then nearest centroid classifier is applied for classification. It should be noted that when $\mathbf{H}$ or $\mathbf{S}_t$ are singular, a regularized version is used. 

\begin{table}
\caption{Results comparison}\label{result2}
\vspace{1.0mm}
\scalebox{0.72}{
\begin{tabular}[ht]{||c|c|c|c|c|c|c||}
\hline
Dataset& BU & KANADE&JAFFE &ORL& YALE&AR \\
\hline
LDA&0.5729&0.6898&0.5571&0.9725&0.9593&0.9688\\
\hline
\cite{Tang2005LinearLDA}&0.5743&0.6857&0.5714&0.9800&0.9564&0.9681\\
\hline
\cite{Jarchi2008ALDA}&0.5957&0.6898&0.5381&0.9800&$\boldsymbol{0.9597}$&0.9692\\
\hline
$SwLDA_{41}$&$0.6500$&$0.7020$&$\boldsymbol{0.5905}$&$\boldsymbol{0.9850}$&$\boldsymbol{0.9597}$&$\boldsymbol{0.9696}$\\
\hline
$SwLDA_{42}$&$\boldsymbol{0.6786}$&$\boldsymbol{0.7224}$&0.5476&0.9450&0.9593&$\boldsymbol{0.9696}$\\
\hline
\end{tabular}}
\end{table}
\section{Experiment Results}
\label{res}
In our experiments, we evaluate the performance of proposed $SwLDA$, traditional LDA and two weighted LDA approaches mentioned in section 2 on six public facial image datasets: BU, KANADE, JAFFE, ORL, YALE and AR. We evaluate the performance of the proposed $SwLDA$ approaches, as illustrated in Table \ref{result1}. The results of $SwLDA_{ij}$ illustrate classification accuracy obtained by using the matrices $\mathbf{S}_t^{(ij)}$ and $\mathbf{S}_b^{(i)}, i \in \{1, 2, 3, 4 \}, j \in \{ 1, 2 \} $. The result of traditional LDA is considered as baseline. The results comparison of baseline, Tang's work \cite{Tang2005LinearLDA}, Jarchi's work \cite{Jarchi2008ALDA} and our work are presented in Table \ref{result2}. We implement standardization on all datasets before training and split each dataset into 5 folds for cross-validation. When obtaining $\mathbf{W}_c$, we select $k$-NN graphs with $k \in \min(5, 0.1*N_{c})$ or fully connected graphs to evaluate its impact on the results. As shown, the best performances over datasets BU and KANADE are both achieved by using $SwLDA_{42}$ with fully connected graphs. $SwLDA_{41}$ is the most effective over dataset JAFFE. The maximal improvement is $10.57\%$ on dataset BU using $SwLDA_{42}$ with fully connected graphs, compared to the result of traditional LDA. That over Tang's work \cite{Tang2005LinearLDA} is $0.14\%$ and over Jarchi's work \cite{Jarchi2008ALDA} is $2.28\%$. $SwLDA_{12}$ and $SwLDA_{22}$ work better than $SwLDA_{32}$ and $SwLDA_{42}$ apparently on datasets JAFFE and ORL. Fully connected graphs works better than $k$-NN graphs does over YALE dataset for all cases. Graph connection does not affect the classification accuracy using $SwLDA_{11}$, $SwLDA_{21}$, $SwLDA_{41}$, $SwLDA_{22}$ and $SwLDA_{32}$ over dataset AR.

\section{conclusion}
\label{sec:con}
In this paper, we propose weighted LDA variants based on a probabilistic definition of visual saliency estimation. We follow a class-specific saliency estimation process in order to determine the contribution of each sample in the optimization problems solved for discriminant subspace learning. Then, we employ our new approaches to six public datasets for evaluation and comparison with related LDA methods. Our new definitions target to reveal connections between each sample in every class, and further solve shortcomings in weighted LDA variants. Experimental results sufficiently demonstrate that the highest classification accuracy is always with one of our proposed approaches over these six facial image datasets.

\bibliographystyle{IEEEbib}
\bibliography{strings,refs}

\end{document}